\documentclass[11pt]{article}

\usepackage[a4paper, margin=1in]{geometry} 
\usepackage{times}
\usepackage{amsmath, amssymb, amsthm}
\usepackage{enumitem}
\usepackage{subcaption}
\usepackage{algorithm}
\usepackage{algorithmicx}
\usepackage{algpseudocode}
\usepackage{csvsimple}
\usepackage{booktabs}
\usepackage{tikz}
\usetikzlibrary{positioning, arrows.meta, shapes.geometric}
\usepackage{float}
\usepackage{newfloat}
\usepackage{listings}
\usepackage{hyperref}
\usepackage{url}

\hypersetup{
    colorlinks=true,
    linkcolor=blue,
    citecolor=blue,
    urlcolor=blue,
    pdfauthor={Your Name},
    pdftitle={Your Paper Title}
}

\newtheorem{theorem}{Theorem}

\newtheorem{assumption}{Assumption}
\newtheorem{remark}{Remark}

\usepackage[numbers,sort&compress]{natbib}  % or omit 'numbers' for (Author et al., Year)
\usepackage{etoolbox}

\makeatletter
\renewcommand{\NAT@nmfmt}[1]{#1}           % plain formatting of names

\makeatother

\title{An Agent-Based Framework for the Automatic Validation of\\ Mathematical Optimization Models}
\author{
  Alexander Zadorojniy\thanks{IBM Research} \\
  IBM Research \\
  \texttt{zalex@il.ibm.com}
  \and
  Segev Wasserkrug \\
  IBM Research \\
  \texttt{segevw@il.ibm.com}
  \and
  Eitan Farchi \\
  IBM Research \\
  \texttt{farchi@il.ibm.com}
}

\date{}

\begin{document}

\maketitle

\begin{abstract}
Recently, using Large Language Models (LLMs) to generate optimization models from natural language descriptions has became increasingly popular. However, a major open question is how to validate that the generated models are correct and satisfy the requirements defined in the natural language description. In this work, we propose a novel agent-based method for automatic validation of optimization models that builds upon and extends methods from software testing to address optimization modeling . This method consists of several agents that initially generate a problem-level testing API, then generate tests utilizing this API, and, lastly, generate mutations specific to the optimization model (a well-known software testing technique assessing the fault detection power of the test suite). In this work, we detail this validation framework and show, through both theory and experiments, the high quality of validation provided by this agent ensemble in terms of the well-known software testing measure called mutation coverage. 
\end{abstract}

\section{Introduction}\label{Intro}
Mathematical optimization is a formal mathematical modeling technique that can be applied to help address many important real-world decision-making problems. Although our approach is applicable to general mathematical‑programming problems, for clarity of exposition in this paper we focus on the subset of optimization problems that can be formulated as linear programs (LPs) of the following form:
\begin{alignat*}{3}
\text{minimize}\;   & c^{\mathsf T}x                &\quad& \text{(objective function)} \\[4pt]
\text{subject to}\; & A x \le b,                    &      & \text{(inequality rows)}   \\[4pt]
                    %& A_{\mathrm{eq}}x = b_{\mathrm{eq}}, & & \text{(equality rows)}    \\[2pt]
                    & \ell \le x \le u              &      & \text{(variable bounds)}
\end{alignat*}

Where $x \in \mathbb{R}^{n}$ is a vector of decision variables, $c \in \mathbb{R}^{n}$ is cost or profit vector, $A \in \mathbb{R}^{m\times n},\; b \in \mathbb{R}^m$ is a matrix and right-hand side for the inequality block \(Ax \le b\), $\ell, u \in (\mathbb{R}\cup\{-\infty,\infty\})^{n}$ is component-wise lower and upper bounds; set \(\ell_i=-\infty\) or \(u_i=\infty\) when a bound is absent. 

Modeling a real world problem using mathematical optimization is a lengthy process, requiring significant skills as it requires modeling a problem using a set of decision variables whose values should be selected so as to optimize some objective function while satisfying a set of constraints. Therefore, in recent years, many works have explored using LLMs to generated mathematical optimization models from natural language descriptions~\cite{OptimusICML,ramamonjison2023nl4opt, xiao2024chainofexperts, li2023large} so as to make optimization modeling more widely accessible and reduce the time required to create such models. Such efforts have met with considerable success,  which has even resulted in some of the commercial software vendors offering LLM based \textit{modeling assistants}~\cite{gurobi2024ai}. However, in spite of this success there are often still mistakes in the LLM generated models. Therefore, a significant open question is how to automatically validate the correctness of such models.

This issue of generated model validation arises from two characteristics of optimization modeling: the first is the fact that the natural language description must be translated into a formal model, and the second is that many, significantly different model formulations can solve the problem specified in natural language. The reason that there is such a variety of ways to model the problem is that ultimately, two formulations are suitable for solving the same problem if they capture the same set of optimal solutions (a solution is optimal if it is a solution that maximizes/minimizes the value of the objective function while satisfying the constraints). To see a concrete example of such two different valid models, consider the following decision problem: A mid-sized precision-engineering plant manufactures two profitable items during each eight-hour shift - heavy-duty gearbox housings and precision mounting brackets. Every housing and every bracket requires one hour in total for deburring, gauging, and final sign-off at assembly and inspection benches. A housing occupies the machine and finishing cell for two hours, whereas a bracket needs one hour. The combined line is capped at ten machine-hours each day. Each gearbox housing results in 120\$ net profit per unit and each mounting bracket in  90\$  net profit per unit. The problem is to decide how many housing units ($x$) and how many brackets ($y$) to produce so as to optimize the profit while satisfying the time constraints of the plant.
One possible formulation of the problem is:
% \begin{subequations}\label{eq:lp}
% \begin{align}
% \text{maximize}\quad & 120\,x + 90\,y \label{eq:lp_obj}\\[4pt]
% \text{subject to}\quad
% & x + y \le 8, \label{eq:lp_c1}\\[2pt]
% & 2x + y \le 10, \label{eq:lp_c2}\\[2pt]
% & x \ge 0,\; y \ge 0. \label{eq:lp_bounds}
% \end{align}
% \end{subequations}
\begin{subequations}\label{eq:lp}
\begin{align}
\text{maximize}\quad & 120\,x + 90\,y \label{eq:lp_obj} \\[4pt]
\text{subject to}\quad
& x + y \le 8, \quad
  2x + y \le 10, \quad
  x, y \ge 0 \label{eq:lp_bounds}
\end{align}
\end{subequations}

An equivalent formulation of the problem is one in which the same set of constraints are used but  the objective function is scaled to a different currency resulting in the objective $\textbf{maximize} \quad  1200\,x \;+\; 900\,y$. These two correct formulations will result in two different optimal objective function values, even though they will both provide the same optimal solutions in terms of the $x$ and $y$ values. Another equivalent optimization model is the \textit{dual formulation}~\cite{bertsimas1997introduction} which again results in an optimal solution to the original problem but with a very different formulation (equations \eqref{eq:dual_obj}-\eqref{eq:dual_bounds} show the dual formulation for the problem described in equations \eqref{eq:lp_obj}-\eqref{eq:lp_bounds}): 

\begin{subequations}\label{eq:dual}
\begin{align}
\text{minimize}\quad & 8\,u_1 + 10\,u_2 \label{eq:dual_obj} \\[4pt]
\text{subject to}\quad
& u_1 + 2u_2 \ge 120, \quad
  u_1 + u_2 \ge 90, \quad
  u_1, u_2 \ge 0 \label{eq:dual_bounds}
\end{align}
\end{subequations}

%%%%%%%%%%
Another reason for the use of different models for the same problem stems from the fact that once modeled in mathematical form, mathematical optimization models are often solved by generic \textbf{Optimization engines} such as
IBM\,ILOG\,CPLEX\,Optimization\,Studio~\cite{IBM_ILOG_CPLEX_2025}
or the Gurobi\,Optimizer~\cite{Gurobi_2025}, and different formulations could have a very significant impact on the time it takes such engines to find optimal, or even good, solutions. This results in additional incentives to create different, more efficient models, for the same problem. 

Irrespective of the underlying reason, the existence of different equivalent models may mean that it is quite hard to understand whether any given model is indeed a model that provides an optimal or even a feasible solution for a business problem. This makes automatic validation of  generated models especially challenging - even when a ground-truth model is available.
Existing techniques for automatic model validation include testing syntactic equivalence between the generated and ground truth models ~\cite{ramamonjison2023nl4opt} and comparing only their optimal objective values~\cite{OptimusICML}. Both techniques are unreliable: both would fail, for example when two optimization models differ by the scaling of the objective function as described above, resulting in a false negative. The use of objective value comparison could also result in a false positive, as can be seen by considering the case illustrated in Figure~\ref{fig:sub3}. This figure  contrasts a correct formulation with an incorrect one that satisfies only a subset of feasible scenarios. If we evaluate the models on a default instance that the flawed formulation happens to satisfy, the comparison of objective values alone will yield a false positive. Consequently, relying exclusively on objective-value checks is insufficient for robust model validation. 

\paragraph{Contribution} To address this challenge, we propose a novel agent-based approach that builds upon best practices in software testing for validating optimization models. Our approach is an automated one, and includes a set of several LLM based agents that utilize the natural language description of the problem. These agents include: an agent that uses an LLM to create a problem level, rather than an optimization model level, testing interface - enabling the output of the model to be validated against the natural language specification of the problem in a model agnostic manner; an LLM agent to create unit tests - intended to validate model correctness by utilizing the solutions provided by the model through the business level interface; an LLM that generates an auxiliary optimization model aimed at providing feedback on the correctness of unit tests; and an LLM agent to generate optimization modeling specific  \textit{mutations} - a well-known software testing concept whose goal is to help assess the fault-detecting power of a test suite. We executed the test suite on the auxiliary optimization models deterministically, guaranteeing that every test yields a definitive Pass or Fail result. In the rest of this work we detail our approach, and demonstrate its effectiveness through empirical validation.

\subsection{Related Work}

\paragraph{Optimization Modeling and Verification using LLMs}
\cite{xiao2024chainofexperts} propose a multi-agent LLM system that automatically derives optimization models from textual problem descriptions. Validation was carried out by a group of experts who wrote at least five tests per problem to verify code correctness. This strategy is labor-intensive, and a suite of only five tests is generally inadequate — even for academic benchmark datasets. For automatic validation, their workflow depends on an additional LLM-based agent to review model correctness; the agent’s feedback is then fed to the other agents to steer subsequent iterations. Because LLMs are susceptible to hallucination \cite{Huang_2025}, this review loop is inherently error-prone.

\cite{OptimusICML} present an LLM-driven, multi-agent system that automatically constructs optimization models from natural-language problem descriptions. Although this is an important step toward fully automated model generation, their validation strategy compares only the optimal objective values of the generated models with manually computed reference values. As described in Section~\ref{Intro}, agreement (resp. disagreement) on the objective value alone does not guarantee that a model is correct (resp. incorrect). Moreover, the true optimum is frequently unknown apriori — making this form of validation unreliable.

\cite{li2023large} \emph{proposes} a system that answers users’ supply-chain queries (e.g., \emph{what-if} analyses).  
The authors validated correctness by manually creating five scenarios with known optimal solutions.  
For each scenario, they automatically generated question-and-answer sets—for example, ``What if we prohibit shipping from supplier~X to roaster~Y?'' or ``Demand at café~Z increases by 10\%.''  
Macros substitute random entity names, yielding thousands of distinct yet structurally similar questions.  
Although this approach provides valuable validation for the supply-chain-specific scenarios described in this work, crafting such templates for every problem is impractical, and optimal solutions are often unknown or not unique.

\paragraph{Software Testing}
Traditional software testing ensures that code behaves as expected by executing test cases that uncover defects, validate functionality, and exercise edge cases. Recently, large language models (LLMs) have been applied to automate and enhance this process. For example, Meta’s TestGen-LLM tool~\cite{alshahwan2024automated} automatically generates functional tests for conventional software.

Our work, by contrast, does not validate traditional software; instead, it targets the creation and verification of mathematical programming models. We nevertheless highlight recent advances in LLM-based test generation for completeness.

To improve test effectiveness and guarantee sufficient \textit{test coverage}, a variety of coverage approaches have been proposed in the software testing domain ~\cite{10.1093/comjnl/bxm021}. One widely adopted coverage approach is mutation testing, which systematically creates small program variants (mutants) to assess the fault-detecting power of a test suite~\cite{mutation}. As a part of our work presented in this paper, we explore how the mutation-testing technique can be transferred to the domain of mathematical programming, using LLMs to automate mutant generation and evaluation.

\subsection{Background - Mutation Testing}
Mutation testing \cite{PAPADAKIS2019275} is a technique used to evaluate the effectiveness of a test suite.   It captures the intuitions that some categories of software bugs occur as a result of small perturbations to the software code, e.g., writing $\leq$ instead of $\geq$ or visa versa.  Such changes to the code are called mutations.  In more details the method works as follows.

\begin{enumerate}
    \item \textbf{Mutations introduction}: Small changes, called mutants, are intentionally injected into the program's source code. These mutations might involve altering operators (e.g., changing + to -), modifying constants, or tweaking control structures.
    \item \textbf{Test Execution}: The entire test suite is executed against the program being tested  and its mutants. The idea is to simulate potential faults that could occur in the code.  Note that the fault could be in the original program as it should have had $\geq$ in a condition but had $\leq$ in a condition. Introducing a mutant that changes $\geq$ to $\leq$ is aimed at revealing that programming error. 
    \item \textbf{Tests effectiveness evaluation}: If a mutant causes the test suite to fail, it is considered "killed," indicating that the tests were effective in catching that change. If a mutant does not trigger any test failures, it "survives." A surviving mutant suggests that there might be a gap in the test suite, as it failed to detect a potential defect.
    \item \textbf{Tests coverage improvement}. By identifying surviving mutants, developers can write additional tests or refine existing ones and ensure that the test suite is strong enough to catch subtle programming errors.
\end{enumerate}

Common type of mutations include value and decision mutations. Value mutations involve changing the literal or constant values in the code. These mutations simulate scenarios where an incorrect value is used, helping ensure that the tests validate the proper handling of data. Decision mutations focus on the logical structure of the code, especially the conditions that control the flow of execution (such as if or while statements). These mutations simulate errors in decision-making logic.

\section{Tests Coverage for Optimization Models}

Software coverage \cite{10.1093/comjnl/bxm021} is aimed at giving a measure of the adequacy of testing.  For example, it may count the percentage of statements that were executed by any test in the testing set.  
Compared to software development, optimization problems typically involve fewer programming constructs, often just a few dozen instead of thousands, and those requirements tend to be more clearly defined, change less frequently, and involve less integration with existing systems. However, optimization problems are more challenging when it comes to the mathematical modeling, the choice of algorithms and solvers, and managing computational complexity. Thus, we expect that testing and validation of optimization problems will be generally simpler than it is for large-scale software projects. For testing of the optimization models we will use decision and value mutations which should be sufficient to cover most frequent automatically generated optimization models issues. 
\subsection{Mutation Coverage of Optimization Models}
As described above, a common way of defining mutation coverage is through the ratio of number of "killed" mutations to overall mutations generated.  In our context mutations are applied to the constraints of a given problem in the benchmark.   For example, a constraint of a given problem in the benchmark may state that $x + y \leq 8$ and its mutant may be $x + y \leq 7$ (changing the constant 8 to the constant 7).

In addition, the concept of a \emph{killed mutation} applies to our context.  
A mutant is said to be \emph{killed} if we can demonstrate that the mutated mathematical program fails to pass the tests intended to validate that the model solves the optimization problem stated in natural language.  
Let \(K\) denote the number of killed mutants for a dataset and \(M\) the total number of mutants for that dataset.  
The \textit{mutation coverage} for the dataset (\(\widehat{MC}\)) is then defined as
\begin{equation}
\widehat{MC}\,[\%] = \frac{K}{M} \times 100\,\%.
\label{eq:mutation_coverage}
\end{equation}
Throughout this paper, we define coverage as mutation coverage.

\begin{figure}
    \centering
    % First subfigure
    \begin{subfigure}[b]{0.3\textwidth}
        \centering
        \includegraphics[width=\textwidth]{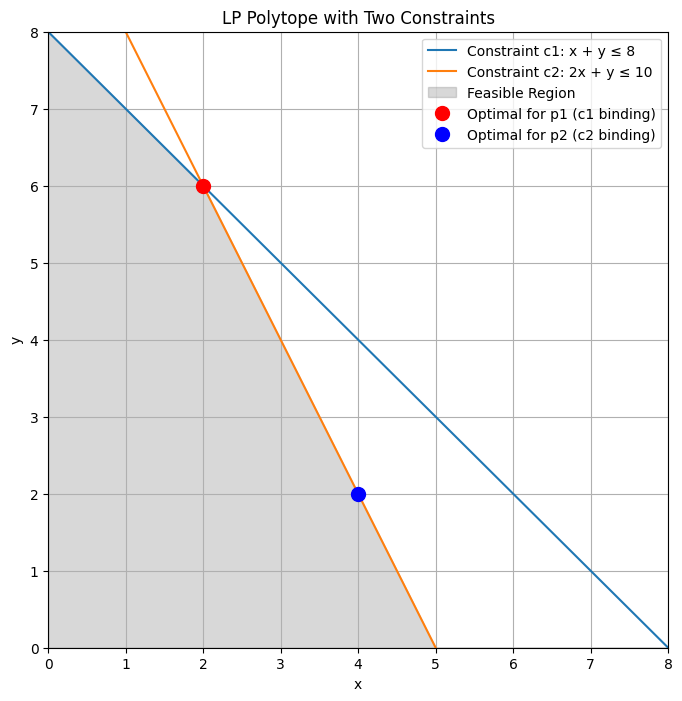}
        \caption{LP - non-binding \\ solution.}
        \label{fig:sub1}
    \end{subfigure}
    \hfill
    % Second subfigure
    \begin{subfigure}[b]{0.3\textwidth}
        \centering
        \includegraphics[width=\textwidth]{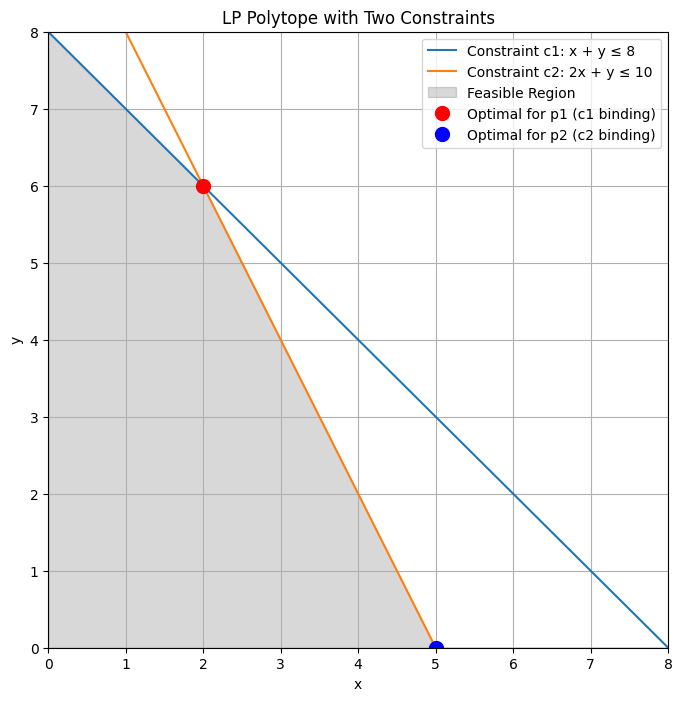}
        \caption{LP - non-binding optimal solution.}
        \label{fig:sub2}
    \end{subfigure}
    \hfill
    % Third subfigure
    \begin{subfigure}[b]{0.3\textwidth}
        \centering
        \includegraphics[width=\textwidth]{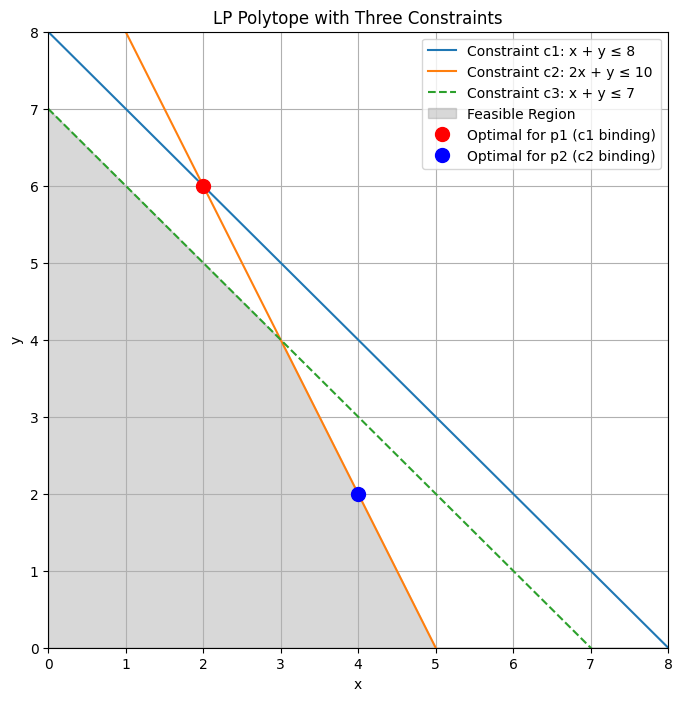}
        \caption{LP Mutated and Original Problem.}
        \label{fig:sub3}
    \end{subfigure}
    \caption{LP - What we want to cover.}
    \label{fig:three_side_by_side}
\end{figure}

To illustrate the importance of mutation coverage, consider the same daily production-planning problem evaluated with two different profit functions but identical constraints. The original profit function is $120x + 90y$ and the modified one is $140x + 60y$. With the correct set of constraints, both profit functions are supported—each yields an optimal, feasible solution (Figures~\ref{fig:sub1} and~\ref{fig:sub2} respectively). However, if constraint c1 is mistakenly replaced by constraint c3 (Figure~\ref{fig:sub3}), only the second profit function remains supported. 

This is precisely what mutation testing does: it deliberately alters the model to reveal potential errors.

\subsection{Mutation Theoretical Analysis}\label{sec:theory}

In this section we provide a mutation-theoretical analysis showing that single-constraint mutations are the most challenging to detect - implying that test suites that detect such mutations are the most comprehensive. We estimate the quality of empirical mutation coverage and derive detection probability bounds for linear programming problems.

% We establish that single-constraint mutations provide the most discriminative evaluation of test-suite quality. The intuition is straightforward: mutating multiple constraints simultaneously increases the probability of detection, making it easier for even weak test suites to kill mutants. Therefore, single-constraint mutations provide a more stringent quality metric.

% \begin{assumption}
% \label{as:local-detect}
% Consider a problem instance $P$ and test suite $T$. Let a mutant be generated by selecting a set $S$ of constraints and applying random perturbations to each constraint in $S$. Then, there exists a constant $q \in (0,1]$ such that for any $s \geq 1$ and any constraint set $S$ with $|S| = s$: if the mutant survives after mutations to some subset $S' \subset S$, then introducing an additional independent mutation to any constraint in $S \setminus S'$ kills the mutant with conditional probability of at least $q$.
% \end{assumption}

\begin{theorem}[Monotonicity of Kill Probability]
\label{thm:s1-hardest}
%Under Assumption~\ref{as:local-detect}, 
For any fixed problem instance $P$ and test suite $T$, the kill probability $p_s(T;P)$ is monotonically nondecreasing in mutation strength $s$:
\[
p_1(T;P) \;\leq\; p_2(T;P) \;\leq\; \cdots \;\leq\; p_s(T;P) \;\leq\; p_{s+1}(T;P) \;\leq\; \cdots
\]
Equivalently, $\forall s \geq 1$:
\[
p_{s+1}(T;P) \;\geq\; p_s(T;P).
\]

\begin{proof}
We construct a strength-$(s+1)$ mutant incrementally:
\begin{enumerate}
    \item Generate the first $s$ constraint mutations according to some distribution $\mu_s(P)$.
    \item Add one additional independently sampled constraint mutation.
\end{enumerate}

Define the following events:
\begin{itemize}
    \item $A_s$: the event that the mutant is killed after the first $s$ mutations
    \item $A_{s+1}$: the event that the mutant is killed after all $s+1$ mutations
\end{itemize}

By the law of total probability:
\[
\Pr(A_{s+1}) = \Pr(A_{s+1} \mid A_s) \cdot \Pr(A_s) + \Pr(A_{s+1} \mid \neg A_s) \cdot \Pr(\neg A_s).
\]

The conditional probability of killing the mutant with the additional mutation, given that it was killed after the first $s$ mutations is $1$.

%By Assumption~\ref{as:local-detect}, 
The conditional probability of killing the mutant with the additional mutation, given that it survived the first $s$ mutations is $\Pr(A_{s+1} \mid
\neg A_s) \;\geq\; \; 0$.
%\neg A_s) \;\geq\; q \;>\; 0$.

% Therefore:
% \[
% \Pr(A_{s+1}) \;\geq\; \Pr(A_s) + q \cdot \Pr(\neg A_s) \;\geq\; \Pr(A_s),
% \]
% which establishes $p_{s+1}(T;P) \geq p_s(T;P)$.

Therefore:
\[
\Pr(A_{s+1}) \;\geq\; \Pr(A_s) + \Pr(\neg A_s) \;\geq\; \Pr(A_s),
\]
which establishes $p_{s+1}(T;P) \geq p_s(T;P)$.

By induction, this inequality holds for all $s \geq 1$, proving that $p_s(T;P)$ is nondecreasing in $s$.
\end{proof}

\noindent\textbf{Corollary.} Single-constraint mutants ($s=1$) provide the most challenging test of test-suite quality, as they yield the minimum kill probability.
\end{theorem}

\begin{proof}
From Theorem~\ref{thm:s1-hardest}, $s=1$ yields the minimum kill probability, making single-constraint mutants the most challenging test for any given test suite.
\end{proof}

%\subsection{Lower Bound from Binding Constraints}

We now quantify detection probability in the single-mutation regime using the structure of optimal solutions of LP.

Consider LP:
\[
\max\{c^\top x : Ax \le b,\; x\in \mathbb{R}^n\}
\]
with $m$ constraints. Let $x^\star$ be an optimal basic feasible solution with binding constraint set:
\[
B := \{\, i\in\{1,\dots,m\} : a_i^\top x^\star = b_i \,\}, \quad d:=|B|.
\]

% Consider a linear program:
% \[
% \max\{c^\top x : Ax \le b,\; x\in \mathbb{R}^n\}
% \]
% where $A \in \mathbb{R}^{m \times n}$ has $m$ constraints and $n$ variables. Let $x^\star$ be an optimal basic feasible solution with binding constraint set:
% \[
% B := \{\, i\in\{1,\dots,m\} : a_i^\top x^\star = b_i \,\}, \quad d:=|B|.
% \]
Note that $d \le m$ is the number of binding constraints (for non-degenerate basic feasible solutions, $d=n$).

\begin{assumption}[Binding-Constraint Detectability]
\label{as:binding-detect}
The test suite $T$ verifies that the mutated model preserves $x^\star$ and/or the optimal objective value. Moreover, mutating any binding constraint $i\in B$ causes $x^\star$ to become infeasible.
\end{assumption}

\begin{theorem}[Detection Probability Lower Bound]
\label{thm:binding-single}
Under Assumption~\ref{as:binding-detect}, a uniformly random single-constraint mutation is detected with probability at least $d/m$:
\[
\Pr(\text{kill}) \ge \frac{d}{m}.
\]
\end{theorem}

\begin{proof}
A uniform random mutation selects each constraint with probability $1/m$. Thus:
\[
\Pr(\text{kill})
\ge
\Pr(\text{binding constraint selected})
=
\frac{d}{m}. \qedhere
\]
\end{proof}

\begin{theorem}[Detection Probability Lower Bound for $s$ Mutations]
\label{thm:binding-multiple}
Under Assumption~\ref{as:binding-detect}, when $s$ constraints are selected uniformly at random without replacement for mutation from the $m$ total constraints (where $1 \le s \le m$), at least one mutation is detected with probability at least:
\[
\Pr(\text{kill}) \ge 1 - \left(1 - \frac{d}{m}\right)^s.
\]
\end{theorem}

\begin{proof}
Consider $s$ constraints selected uniformly at random without replacement from the $m$ total constraints. The probability that \emph{none} of the $s$ selected constraints is binding equals the probability of selecting all $s$ constraints from the $m-d$ non-binding constraints:
\[
\Pr(\text{no binding constraint selected})
=
\frac{\binom{m-d}{s}}{\binom{m}{s}}
=
\prod_{i=0}^{s-1} \frac{m-d-i}{m-i}.
\]

Each factor satisfies $\frac{m-d-i}{m-i} \le \frac{m-d}{m} = 1 - \frac{d}{m}$ for $i \ge 1$ (with equality only at $i=0$), since removing non-binding constraints from the pool decreases the proportion of non-binding constraints in subsequent draws. Therefore:
\[
\Pr(\text{no binding constraint selected})
\le
\left(1 - \frac{d}{m}\right)^s,
\]
where the right-hand side corresponds to sampling with replacement.

Therefore, the probability that at least one binding constraint is selected (and thus detected) is:
\[
\Pr(\text{kill})
=
1 - \Pr(\text{no binding constraint selected})
\ge
1 - \left(1 - \frac{d}{m}\right)^s. \qedhere
\]
\end{proof}

\begin{remark}
For $d \ge 1$, the detection probability is strictly increasing in $s$:
\begin{itemize}
\item For $s=1$: $\Pr(\text{kill}) \ge d/m$ (lower bound, recovers Theorem~\ref{thm:binding-single})
\item For $s \ge m-d+1$: $\Pr(\text{kill}) = 1$ (upper bound achieved)
\item In particular, for $s = m$: $\Pr(\text{kill}) = 1$
\end{itemize}
\end{remark}

\begin{theorem}[Mutation Coverage Estimator]
\label{thm:bench-concentration}
Let $\{X_{i,j} \in \{0,1\} \}$ be independent Bernoulli random variables indexed by problem $i \in \{1,\ldots,N\}$ and trial $j \in \{1,\ldots,r\}$, where $X_{i,j}$ indicates whether the $j$-th mutant of problem $i$ is killed. Define:
\begin{itemize}
    \item $M := Nr$ (total number of mutation trials across all problems)
    \item $\widehat{MC} := \frac{1}{M}\sum_{i=1}^{N}\sum_{j=1}^{r} X_{i,j}$ (empirical mutation coverage)
    \item $MC :=  \frac{1}{M}\sum_{i=1}^{N}\sum_{j=1}^{r} \mathbb{E}[{X_{i,j}}]$ (true mutation coverage)
\end{itemize}

Then for any $\varepsilon > 0$ and $\delta \in (0,1)$:

\noindent\textbf{(Tail bound)} The tail probability satisfies:
\[
\Pr\Bigl(\,|\widehat{MC} - MC| \;\geq\; \varepsilon\,\Bigr)
\;\leq\; 2\exp\bigl(-2M\varepsilon^2\bigr).
\]

\noindent\textbf{(Confidence interval)} With probability at least $1-\delta$:
\[
|\widehat{MC} - MC|
\;\leq\;
\sqrt{\frac{1}{2M}\log\Bigl(\frac{2}{\delta}\Bigr)}
\;=\;
\mathcal{O}\Bigl(\sqrt{\frac{\log(1/\delta)}{M}}\Bigr).
\]
\end{theorem}

\begin{proof}
The empirical mutation coverage $\widehat{MC}$ is the sample mean of $M$ independent random variables $X_{i,j}$. Applying Hoeffding's inequality for bounded random variables:

For any $\varepsilon > 0$,
\[
\Pr\Bigl(\,|\widehat{MC} - MC| \;\geq\; \varepsilon\,\Bigr)
\;\leq\; 2\exp\Bigl(-\frac{2M\varepsilon^2}{(1-0)^2}\Bigr)
\;=\; 2\exp\bigl(-2M\varepsilon^2\bigr).
\]

To obtain the confidence interval, set the right-hand side equal to $\delta$:
\[
2\exp\bigl(-2M\varepsilon^2\bigr) = \delta
\quad\Longrightarrow\quad
\varepsilon = \sqrt{\frac{1}{2M}\log\Bigl(\frac{2}{\delta}\Bigr)}.
\]

This establishes that with probability at least $1-\delta$, the estimation error is bounded by $\mathcal{O}\bigl(\sqrt{\log(1/\delta)/M}\bigr)$, demonstrating that $\widehat{MC}$ concentrates around $MC$ at rate $\mathcal{O}(M^{-1/2})$.
\end{proof}

\section{Automatic Optimization Validation using LLMs}\label{sec:multi-agent}

\begin{figure}[h]
    \centering
    %--- Subfigure A: Flow diagram ------------------------------------
    \begin{subfigure}[b]{0.45\textwidth}
        \centering
        \includegraphics[width=\linewidth, trim=10pt 100pt 0pt 100pt, clip]{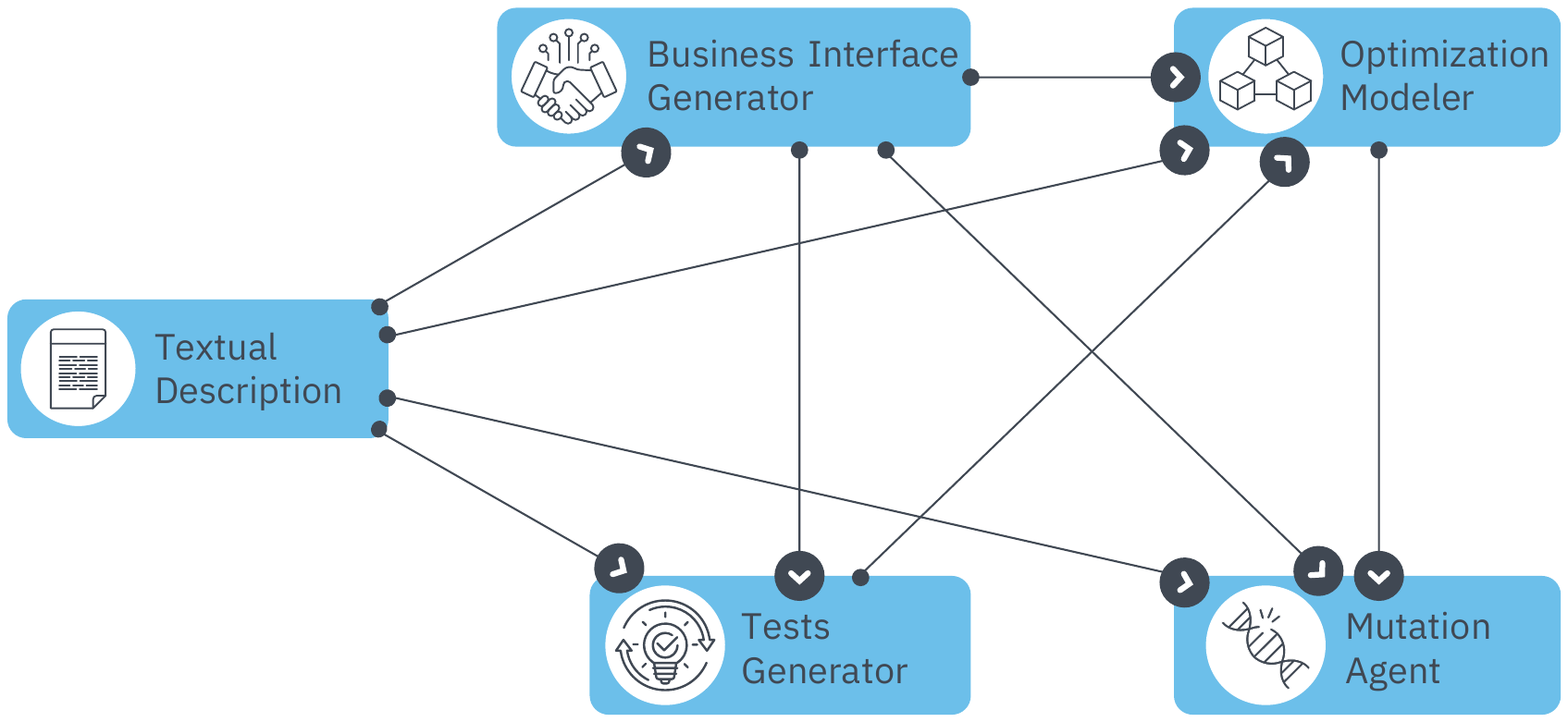}
        \caption{Test Suite Generation Flow}
        \label{fig:flow}
    \end{subfigure}
    \hfill
    %--- Subfigure B: Outcome matrix ----------------------------------
    \begin{subfigure}[b]{0.45\textwidth}
        \centering
        \scriptsize
        \setlength{\tabcolsep}{4pt}
        \renewcommand{\arraystretch}{1.1}
        \resizebox{\linewidth}{!}{%
        \begin{tabular}{|c|c|c|c|c|}
        \hline
        \textbf{Test suite} & \textbf{Model} & \textbf{Mutations} &
        \textbf{Run on model} & \textbf{Run on mutated} \\
        \hline
        Good & Good & Good & Pass & Fail \\
        Good & Good & Bad  & Pass & ? \\
        Good & Bad  & Good & Fail & Likely fail / may pass \\
        Good & Bad  & Bad  & Fail & Likely fail / may pass \\
        Bad  & Good & Good & ?    & ? \\
        Bad  & Good & Bad  & ?    & ? \\
        Bad  & Bad  & Good & ?    & ? \\
        Bad  & Bad  & Bad  & ?    & ? \\
        \hline
        \end{tabular}}
        \caption{Mutation Outcome Matrix}
        \label{fig:mutation_matrix}
    \end{subfigure}

    \caption{Flow and expected outcomes. (a) Test suite generation flow. (b) Outcome matrix for combining test-suite quality (Good/Bad), model correctness (Good/Bad), and mutation validity (Good/Bad): a good suite with a good model should pass on the base model and typically fail on a mutated model; “?” entries indicate cases where outcomes depend on specifics (e.g., weak suites or invalid mutations).}
\end{figure}

\begin{figure}
\centering
    %--- Subfigure A ---------------------------------------------------
    \begin{subfigure}[b]{0.45\textwidth}
    \centering
    \begin{tikzpicture}[
        scale=0.45, transform shape,
        node distance = 12mm and 20mm,
        every node/.style = {font=\small},
        process/.style = {rectangle, rounded corners, draw, align=center,
                          minimum width=3.5cm, minimum height=1.2cm, fill=blue!10},
        io/.style      = {ellipse, draw, align=center,
                          minimum width=3cm,  minimum height=1.0cm, fill=green!10},
        instr/.style   = {rectangle, draw, align=left,
                          minimum width=4cm,  minimum height=1.8cm, fill=yellow!15},
        arrow/.style   = {-{Stealth[length=2mm]}, thick}
    ]
        \node[process] (agent) {Optimization Model\\Generator Agent};
        \node[io,    left=of agent]        (prob)   {Problem\\Description};
        \node[io,    below left=of agent]  (iface)  {Business\\Interface};
        \node[io,    below=of iface]       (tests)  {Unit\\Tests};
        \node[instr, right=of tests]       (guides) {Optimization Model\\Generation Instructions};
        \node[io, right=of agent]          (output) {Generated\\Optimization Model\\Code};
        \draw[arrow] (prob)  -- (agent);
        \draw[arrow] (iface) -- (agent);
        \draw[arrow] (tests) -- (agent);
        \draw[arrow] (guides) -- (agent);
        \draw[arrow] (agent) -- (output);
    \end{tikzpicture}
    \caption{Optimization Model Generator}
    \label{fig:omg_agent}
    \end{subfigure}
    \hfill
    %--- Subfigure B ---------------------------------------------------
    \begin{subfigure}[b]{0.45\textwidth}
    \centering
    \begin{tikzpicture}[
        scale=0.45, transform shape,
        node distance = 12mm and 20mm,
        every node/.style = {font=\small},
        process/.style = {rectangle, rounded corners, draw, align=center,
                          minimum width=3.7cm, minimum height=1.3cm, fill=blue!10},
        io/.style      = {ellipse, draw, align=center,
                          minimum width=3.2cm, minimum height=1.0cm, fill=green!10},
        instr/.style   = {rectangle, draw, align=left,
                          minimum width=4.2cm, minimum height=1.8cm, fill=yellow!15},
        arrow/.style   = {-{Stealth[length=2mm]}, thick}
    ]
        \node[process] (agent) {Mutated\\Optimization\\ModelAgent};
        \node[io,    left=of agent]        (prob)   {Problem\\Description};
        \node[io,    below left=of agent]  (iface)  {Business\\Interface};
        \node[io,    below=of iface]       (base)   {Baseline\\Model};
        \node[instr, right=of base]        (mtype)  {Mutation\\Type / Rules};
        \node[io, right=of agent]          (output) {Mutated\\Optimization\\Model};
        \draw[arrow] (prob)  -- (agent);
        \draw[arrow] (iface) -- (agent);
        \draw[arrow] (base)  -- (agent);
        \draw[arrow] (mtype) -- (agent);
        \draw[arrow] (agent) -- (output);
    \end{tikzpicture}
    \caption{Mutated Optimization Model Agent}
    \label{fig:mut_agent}
    \end{subfigure}

    \caption{Agents and their I/O. (a) Optimization Model Generator — inputs: problem description, business interface, unit tests, and generation instructions; output: optimization model code. (b) Mutated Optimization Model Agent — inputs: problem description, business interface, baseline model, and mutation rules; output: mutated optimization model.}
\end{figure}
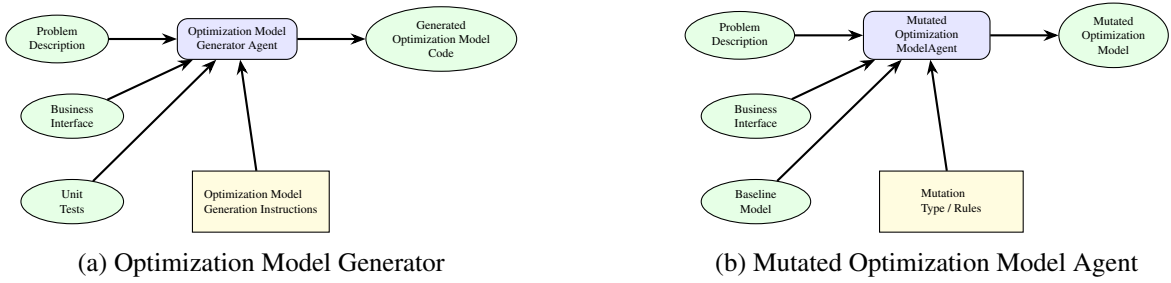

\begin{figure}[H]
    \centering
    %--- Subfigure A ---------------------------------------------------
    \begin{subfigure}[b]{0.45\textwidth}
    \centering
    \includegraphics[width=1.0\textwidth,
                     trim=100pt 100pt 0pt 120pt, clip]{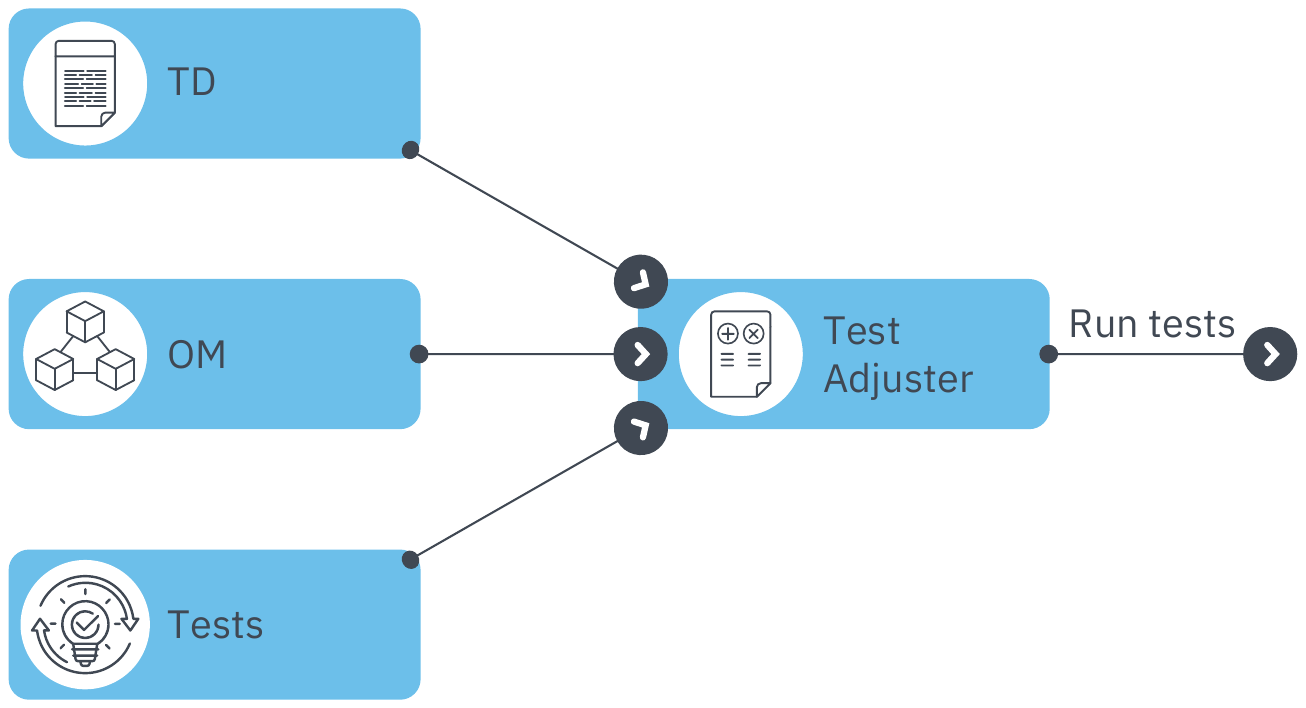}
    \caption{Tests Adjuster}
    \label{fig:agent}
    \end{subfigure}
    \hfill
    %--- Subfigure B ---------------------------------------------------
    \begin{subfigure}[b]{0.45\textwidth}
    \centering
    \begin{tikzpicture}[
        scale=0.45, transform shape,
        node distance = 12mm and 20mm,
        every node/.style = {font=\small},
        process/.style = {rectangle, rounded corners, draw,
                          align=center, minimum width=3.5cm,
                          minimum height=1.2cm, fill=blue!10},
        io/.style      = {ellipse, draw, align=center,
                          minimum width=3cm,  minimum height=1.0cm,
                          fill=green!10},
        instr/.style   = {rectangle, draw, align=left,
                          minimum width=4cm,  minimum height=1.8cm,
                          fill=yellow!15},
        arrow/.style   = {-{Stealth[length=2mm]}, thick}
    ]
        \node[process] (agent) {Test\\AdjusterAgent};
        \node[io,    left=of agent]        (prob)  {Problem\\Description};
        \node[io,    below left=of agent]  (model) {Optimization\\Model};
        \node[io,    below=of model]       (tests) {Original\\Unit Tests};
        \node[io, right=of agent]          (output){Adjusted\\Unit Test\\Suite};
        \draw[arrow] (prob)  -- (agent);
        \draw[arrow] (model) -- (agent);
        \draw[arrow] (tests) -- (agent);
        \draw[arrow] (agent) -- (output);
    \end{tikzpicture}
    \caption{Test Adjuster Agent}
    \label{fig:test_adjuster_agent}
    \end{subfigure}

    \caption{Agents and their I/O. (a) Tests Adjuster — input: problem description, optimization model, and original unit tests; output: adjusted API-aligned test suite. (b) Test Adjuster Agent — same I/O represented as a process diagram.}
    
\end{figure}

\noindent
To close the gap between natural-language specifications and the robust validation of optimization solutions, we defined a framework composed of a workflow of four agents. The workflow (appearing in Figure~\ref{fig:flow}) begins with a \emph{Textual Description} authored by a domain expert.\\
This description is parsed by the \textbf{Business-Interface Generator}, an agent which converts the natural language problem statement into a unified, declarative interface that defines the form of the solution that needs to be returned by the solution of the optimization model. All downstream components can consume this interface, which defines entities, parameters, and KPIs related to the solution.\\
Next, the \textbf{Tests Generator} agent uses both the interface and the natural-language specifications to iteratively assemble a diverse suite of test instances, that characterizes the expected behavior of any subsequent optimization artifact.\\
The \textbf{Optimization Modeler} agent then builds an auxiliary optimization model that returns an optimization solution conforming to this interface (variables, constraints, and objective) which is used to verify the validity of the tests produced by the tests generator.\\
To ensure that the tests generated by the generator provide good testing coverage,  a dedicated \textbf{Mutation Agent} creates target mutations which can be injected into the optimization model, so that the tests can be rerun on the mutated models.\\
Surviving mutants indicate potential weaknesses in the test suites, whereas killed mutants strengthen confidence in both the correctness of the test suite and the generated optimization model.
%tests generator scpecify the exact names on top of BUsiness Interface
For each agent we provide a diagram which describes the flow for that agent (Figures~\ref{fig:big_agent},~\ref{fig:utg_agent},~\ref{fig:omg_agent},~\ref{fig:mut_agent},~\ref{fig:test_adjuster_agent})

\begin{figure}[H]
    \centering
    %--- Subfigure A ---------------------------------------------------
    \begin{subfigure}[b]{0.45\textwidth}
    \centering
    \begin{tikzpicture}[
        scale=0.45, transform shape,
        node distance = 12mm and 20mm,
        every node/.style = {font=\small},
        process/.style = {rectangle, draw, rounded corners, align=center,
                          minimum width=3.5cm, minimum height=1.2cm, fill=blue!10},
        io/.style      = {ellipse, draw, align=center,
                          minimum width=3cm, minimum height=1.0cm, fill=green!10},
        instr/.style   = {rectangle, draw, align=left,
                          minimum width=4cm, minimum height=1.8cm, fill=yellow!15},
        arrow/.style   = {-{Stealth[length=2mm]}, thick}
    ]
        \node[process] (agent) {BusinessInterface\\GeneratorAgent};
        \node[io,    left=of agent]        (prob)   {Problem\\Description};
        \node[instr, below left=of agent]  (templ)  {Interface\\Template};
        \node[instr, below right=of templ] (instrn) {Generation\\Instructions};
        \node[io, right=of agent]          (output) {Generated\\Business\\Interface};
        \draw[arrow] (prob)   -- (agent);
        \draw[arrow] (templ)  -- (agent);
        \draw[arrow] (instrn) -- (agent);
        \draw[arrow] (agent)  -- (output);
    \end{tikzpicture}
    \caption{Business Interface Generator}
    \label{fig:big_agent}
    \end{subfigure}
    \hfill
    %--- Subfigure B ---------------------------------------------------
    \begin{subfigure}[b]{0.45\textwidth}
    \centering
    \begin{tikzpicture}[
        scale=0.45, transform shape,
        node distance = 12mm and 20mm,
        every node/.style = {font=\small},
        process/.style = {rectangle, rounded corners, draw, align=center,
                          minimum width=3.5cm, minimum height=1.2cm, fill=blue!10},
        io/.style      = {ellipse, draw, align=center,
                          minimum width=3cm, minimum height=1.0cm, fill=green!10},
        instr/.style   = {rectangle, draw, align=left,
                          minimum width=4cm, minimum height=1.8cm, fill=yellow!15},
        arrow/.style   = {-{Stealth[length=2mm]}, thick}
    ]
        \node[process] (agent) {UnitTests\\GeneratorAgent};
        \node[io,    left=of agent]        (prob)   {Problem\\Description};
        \node[io,    below left=of agent]  (iface)  {Business\\Interface};
        \node[instr, below right=of iface] (guides) {Test-Generation\\Instructions};
        \node[io, right=of agent]          (output) {Generated\\Unit Test\\Suite};
        \draw[arrow] (prob)  -- (agent);
        \draw[arrow] (iface) -- (agent);
        \draw[arrow] (guides) -- (agent);
        \draw[arrow] (agent) -- (output);
    \end{tikzpicture}
    \caption{Unit Tests Generator}
    \label{fig:utg_agent}
    \end{subfigure}

    \caption{Agents and their I/O. (a) Business Interface Generator — inputs: problem description, interface template, and generation instructions; output: business-interface code. (b) Unit Tests Generator — inputs: problem description, business interface, and test-generation instructions; output: unit-test suite.}
\end{figure}
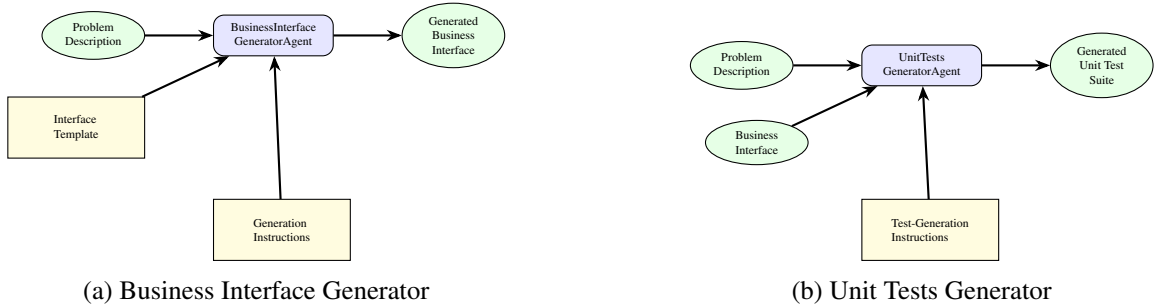

Since LLMs are prone to hallucination~\cite{Huang_2025}, and as we have the agents for the business interface, test suite generation, and model, a single run of the test-suite generation flow may not produce a high-quality outcome, due to errors either one or more of these (to see this consider Table~\ref{fig:mutation_matrix}). It is clear that for generating a good test suite, our goal is to be in the first row of the table - the one in which the originally generated model passes the test suite but the mutated model does not;therefore, we run the flow until the generated model passes the test suite (or up to a maximum number of iterations). Although we do not establish a theoretical upper bound on the required number of iterations, we empirically show that it remains small on the well-known comprehensive optimization model dataset~\cite{udell_nlp4lp_2024}. This iterative execution of the workflow appears in Algorithm~\ref{alg:multiagent}. Note that as the model is not provided as input to the test generator the test suite generation does not depend on the model produced.

%\begin{minipage}{\linewidth}
\begin{algorithm}[H]
\caption{Tests Suite Generation Algorithm.}
\label{alg:multiagent}
\begin{algorithmic}[1]
\Require Textual description $TD$
\Ensure Business interface $BI$, test suite $TS$, optimization model $OM$, mutation artifacts $MA$
\Statex\textit{/* Initialization from the specifications */}
\State $BI \gets \Call{BusinessInterfaceGenerator}{TD}$          %\Comment{BIG : TD 
\Statex\textit{/* Iterative refinement */}
\Repeat
    \State $TS \gets \Call{TestsGenerator}{TD, BI}$          
    \State $OM \gets \Call{OptimizationModeler}{TD, BI, TS}$
    \State Run $TS$ for $OM$
    \State $MA \gets \Call{MutationAgent}{TD, BI, TS, OM}$       
\Until{$TS$ and $OM$ either Pass or Number of Iterations Outed} \Comment{MA is a byproduct}
\State \Return $(BI, TS, OM, MA)$
\end{algorithmic}
\end{algorithm}

\noindent\textbf{Algorithm~\ref{alg:multiagent}} begins with a natural-language specification $TD$ and first extracts a high-level business interface $BI$ (line 1). It then enters an iterative loop (line 2) in which (line 3) a test suite $TS$ is regenerated from both $TD$ and the current $BI$, (line 4) an optimization model $OM$ is (re)constructed to satisfy $TD$, $BI$, and the new tests, (line 5) the tests are executed against $OM$, and (line 6) a mutation agent produces artifacts $MA$ that stress-test the system. The loop terminates when the model passes all tests (or a number of iterations budget elapses), after which the algorithm returns the four artifacts $,(BI,TS,OM,MA),$ with $MA$ serving as a useful by-product for future robustness checks.

Note that Algorithm~\ref{alg:multiagent} operates similarly to a "Monte Carlo Search", as it does not use feedback from failures in previous iterations. This is based on our observation that such feedback does not improve LLM performance for the experiments we did; however, it may still prove useful for more complex problems, in which case it could be incorporated as an automatic prompt-adjustment mechanism for the agents.

\section{Applying Existing Test Suites to Legacy Optimization Models}
The test-suite generator produces an auxiliary optimization model. Ultimately, however, we want to enable users to test optimization models that were created outside of this framework. To accommodate this, we developed an additional agent that adapts the generated test suite to any existing optimization model and executes the tests against it.

The figure illustrates a streamlined \emph{test-adjustment workflow} that utilized this agent.  
On the left reside the three artifacts available: the natural-language specification (\textbf{TD}), the external optimization model (\textbf{OM}) to be tested, and a regression test-suite (\textbf{Tests}).  
These artifacts are simultaneously fed into the \textbf{Test Adjuster} agent, whose task is to reconcile the suite with both the narrative description and the mismatched assertions related to API, updating parameters and their names.
The resulting, consistently updated suite is executed against the optimization model.  
\begin{algorithm}
\caption{LLM based Adjustments of the Test Suite.}

\label{alg:adjuster}
\begin{algorithmic}[2]
\Require Textual description $TD$, Optimization Model $OM$, Tests.
\State $AT \gets \Call{TestsAdjuster}{TD, OM, Tests}$
\State \Return $AT$
\end{algorithmic}
\end{algorithm}

\noindent\textbf{Algorithm~\ref{alg:adjuster}} receives the textual specification $TD$, optimization model $OM$, and a pre-existing test suite.  It invokes a large-language-model agent, \emph{TestsAdjuster}, which cross-checks $TD$ against the behavior of $OM$ and revises the tests accordingly, producing a new artifact $AT$. The adjusted test suite $AT$ is returned for subsequent validation cycles, ensuring that the evolving model remains aligned with both the specification and the optimization model.

\section{Experiments}
\paragraph{Data}
To assess the proposed approach, we drew a random sample of
$100$ problems from the \textsc{NLP4LP} benchmark (three problems —specifically 89, 124, and 269 — were excluded because their descriptions were ambiguous).~\footnote{All problem instances, source code, and prompts are provided in the supplementary material. The prompts were tuned using problems that are not part of the selected set of 100 problems.} 

The NLP4LP data set consists of problems described as text (e.g., a basic scheduling or resource allocation task), constraints (e.g., limits such as capacity or budget constraints) and a goal (e.g., minimizing cost or maximizing profit). Each problem's folder is organized in a modular way so that each problem instance is packaged with several files that separate its different aspects. For our experimentation we used 
description.txt for problems description and solution.json for solution comparison for default scenarios.  

\paragraph{Computational Experiments - Tests Suite Quality Validation}
The tests for each problem were generated using our workflow with two different large-language-model (LLM) configurations:
\begin{itemize}
  \item all agents using the \texttt{o1-preview} LLM; and
  \item a hybrid setup that combines \texttt{gpt-4o} for the auxiliary optimization model generator, and  \texttt{o1-preview} for all other agents (the goal of this was to test the impact of a less powerful auxiliary optimization model generator on the validity of the generated test suite).
\end{itemize}
In both cases, the temperature was fixed at $T = 1.0$. In the hybrid configuration, \texttt{gpt-4o} LLM was used exclusively for the OM agent. 
%Since a temperature of \( T = 1.0 \) introduces non-deterministic behavior in the LLMs, 
We ran\footnote{LLM access was provided via the Azure OpenAI service. Pre- and post-processing were performed on a machine with an 11th Gen Intel(R) Core(TM) i7-11850H CPU @ 2.50GHz and 64GB RAM.} each configuration twice for each problem in the selected dataset, and the entire experimental study took several hours to complete.\footnote{All models, test suites, and their interpretations are provided in the supplementary material.}

Our objective is to obtain a reliable \emph{benchmark-level} estimate of test suite quality across a diverse collection of problems. This aggregate measure provides a robust assessment of a test generation method's overall effectiveness.
By aggregating results over $N \approx 100$ independent problem instances, each contributing $r=2$ independent mutation trials, we obtain $M = Nr \approx 200$ independent Bernoulli observations. Theorem~\ref{thm:bench-concentration} guarantees that this sample size is sufficient for the empirical benchmark-level mutation coverage to concentrate around its expectation, with deviation bounded by $\mathcal{O}(\sqrt{\log(1/\delta)/M})$.

\begin{figure}
    \centering
    %--- Subfigure A: Kill ratio table -------------------------------
    \begin{subfigure}[b]{0.45\textwidth}
        \centering
        \scriptsize
        \setlength{\tabcolsep}{5pt}
        \renewcommand{\arraystretch}{1.1}
        \begin{tabular}{lrrr}
        \toprule
        Model & KILLED & NO & Ratio \\
        \midrule
        all o1-preview    & 142 & 46 & 0.76 \\
        o1-preview + gpt-4o & 131 & 58 & 0.69 \\
        \bottomrule
        \end{tabular}
        \caption{Mutation Kill Ratios}
        \label{tab:kill_ratio}
    \end{subfigure}
    \hfill
    %--- Subfigure B: Iteration distribution plot --------------------
    \begin{subfigure}[b]{0.45\textwidth}
        \centering
        \includegraphics[width=\linewidth]{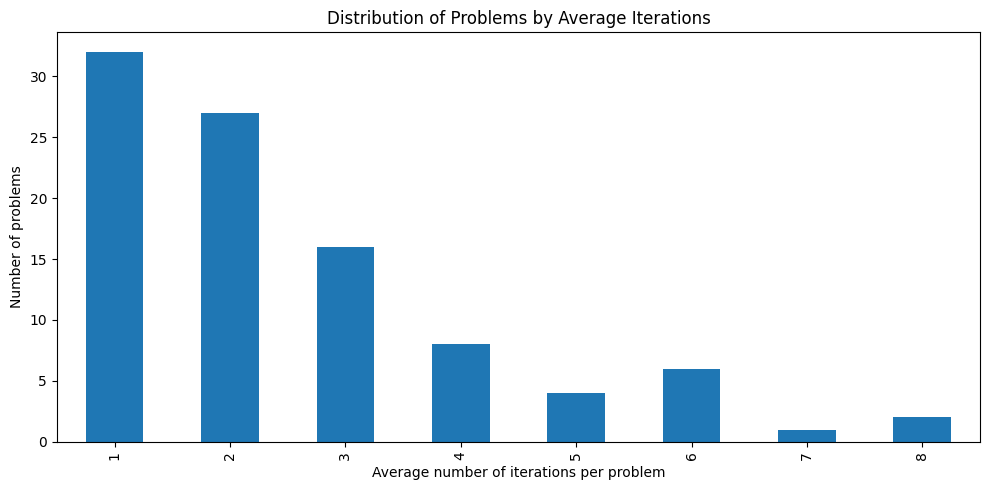}
        \caption{Iteration Distribution}
        \label{fig:avg_distr}
    \end{subfigure}

    \caption{Mutation analysis and convergence. (a) Mutation kill ratios across models: \texttt{o1-preview} achieves a higher ratio (0.76) compared to the combined \texttt{o1-preview + gpt-4o} (0.69). (b) Distribution of the average number of iterations required for convergence, showing most problems converge within a few iterations, with a heavy right-skewed tail.}
\end{figure}

Table~\ref{tab:kill_ratio} summarizes the effectiveness of the mutation-testing campaign for two language-model configurations. 
Across both the weaker (joint \texttt{gpt-4o} and \texttt{o1-preview}) and stronger (\texttt{o1-preview} for all agents) configuration, mutation coverage remained high—at least $69\%$—despite injecting only a single mutation per problem. Note that following theorem~\ref{thm:s1-hardest} a single mutation is designed to mimic the most challenge case.
%difficult and frequent type of error to detect.
% We anticipate higher kill ratios when more (and more diverse) mutations are introduced. 
As expected, using a stronger LLM for the auxiliary optimization model generator resulted in a higher quality set of tests as evidenced by the higher kill ration (76\% as oppose to 69\%).

Another important characteristic evaluated in our framework is the number of iterations required to converge to a ``good'' test suite.  
Figure~\ref{fig:avg_distr} presents a histogram of the \emph{average iterations-to-convergence} recorded for each optimization problem in the benchmark. The distribution is clearly right-skewed.

The vast majority of instances converge quickly: over three-quarters (\(\approx 76\%\)) of the problems require no more than \(3.5\) iterations on average, with the most frequent class centered at \(2.0\) iterations—corresponding to approximately 18 problems. A diminishing tail extends to higher iteration counts. Only a small fraction (around \(10\%\)) of problems require five or more iterations, and the most demanding outlier averages \(8.5\) iterations.

Across both configurations, the \texttt{gpt-4o} variant required, on average, just under three iterations ($2.8$) to converge, whereas the \texttt{o1-preview} variant converged in approximately $2.5$ iterations. To assess the statistical properties of the difference in iterations per problem, we applied the Shapiro--Wilk~\cite{shapiro1965analysis} test to the vector of pairwise differences between configurations. The resulting p-value was very small (\(2.17 \times 10^{-6}\)), indicating that the distribution of differences deviates significantly from normality. Consequently, we employed the Wilcoxon signed-rank test~\cite{conover1999practical} to evaluate whether the difference between the two configurations is statistically significant. The test yielded a p-value of approximately $0.4$, suggesting that the observed difference is not statistically significant.

\paragraph{Auxiliary Optimization Model Validation}
As can be seen from the high kill ratio, most auxiliary optimization models were classified by our system as correct. This outcome supports the idea that small changes to a correct model are effectively detected by our mutation process. To further validate these results, we also tested the correctness of the auxiliary models created by the optimization generator agent using an alternative method.
Specifically, we evaluated whether each generated model, when provided with the input parameters of its corresponding problem instance from the optimization dataset, produced the correct objective function value. This evaluation was conducted across all models generated in all iterations.
If the correct reference value was reproduced in the final iteration, we labeled the auxiliary model for that problem as \textbf{correct}.

For 66 problems, the reference value was matched in all four runs.
For another 12 problems, the reference value was matched in exactly three runs, which led on average to nine auxiliary models per problem being classified as correct.
For the remaining cases where the reference value was reproduced in only 0, 1, or 2 runs, we manually inspected both the auxiliary models and the published reference solutions. Based on this inspection, we identified, on average, 12 additional auxiliary models as correct.
In total, \(87/97\) auxiliary models were deemed correct, corresponding to approximately \(90\%\) accuracy in the auxiliary optimization models generated. This further attests to the validity of the test suites generated by our framework, which consistently classified original optimization models as valid and their mutated variants as invalid with a high kill ratio.

\subsection{Computational Experiments - Testing external optimization models}
To evaluate the usefulness of our testing framework on optimization models generated externally, we required models outside our own pipeline. The dataset we used included only Gurobi solver Python code generated by the Optimus system~\cite{OptimusICML}, which was not suitable for our purposes, as we did not use the Gurobi solver (we used the CPLEX solver instead). In our work, we relied on the \texttt{Pyomo} Python package, as it supports multiple solvers rather than being limited to a specific one.

We therefore evaluated the testing of external models on nine problems from the dataset using the following procedure: we ran the test suites produced by our workflow using only the \texttt{o1-preview} model on optimization models generated for the same problems by the joint \texttt{o1-preview} + \texttt{gpt-4o} pipeline. These models served as external inputs to the end-to-end \texttt{o1-preview} workflow.
We selected nine problems (9, 20, 49, 95, 99, 105, 158, 191, and 199) for which the reference solutions in the NLP4LP dataset differed from those produced by the joint \texttt{o1-preview} + \texttt{gpt-4o} models. These cases were particularly valuable for cross-checking, as the test suites from the two workflows (\texttt{o1-preview} alone vs. joint \texttt{o1-preview} + \texttt{gpt-4o}) produced different results with respect to the reference solutions in the NLP4LP dataset.
Despite mismatched interfaces between models and across iterations (although interfaces remained consistent within a single LLM model and iteration), the adjuster component successfully adjusted every test suite API. This allowed each model generated by \texttt{gpt-4o} to be executed and evaluated against its intended requirements using the test suite generated by the \texttt{o1-preview} model.

\begin{table}[H]
    \centering
    \caption{Tests suite results generated by all $o1-preview$ LLMs for combination of $o1-preview$ and $gpt-4o$ optimization models.}
    \label{tab:comparison_summary}
    \resizebox{0.45\textwidth}{!}{%
        \begin{tabular}{@{}r l p{6cm}@{}}
            \toprule
            \textbf{Problem} & \textbf{Result} & \textbf{Description} \\
            \midrule
             9   & FAIL & \text{Correct optimization model}  \\
            20   & FAIL & \text{Wrong optimization model}  \\
            49   & PASS & \text{Correct optimization model}  \\
            95   & PASS   & \text{Correct optimization model}  \\
            99   & FAIL & \text{Wrong optimization model}  \\
           105   & PASS & \text{Correct optimization model}  \\
           158   & PASS   & \text{Correct optimization model}  \\
           191   & FAIL & \text{Correct optimization model}  \\
           199   & PASS   & \text{Correct optimization model}  \\
            \bottomrule
        \end{tabular}%
    }
\end{table}

We then examined each of the nine benchmark problems listed in Table~\ref{tab:comparison_summary} manually, and carried out additional  targeted LLM checks, to determine whether its optimization model is \emph{Correct} or \emph{Incorrect}. Seven models were found to be correct, while two (i.e., problem 99 and problem 20) exhibited errors: one inequality constraint is written in the wrong direction, which can make the model infeasible or drive it to a sub‑optimal solution, in another one the decision variables are declared as non‑negative reals, but they should be integers; the domain is therefore too relaxed.
To understand the results, refer again to Table~\ref{fig:mutation_matrix} . As to generate the tests we used the  iterative procedure outlined in Section~\ref{sec:multi-agent}, we assume that the tests are good with a high likelihood (i.e., the desired results are in rows 3 and 4 of the table). We expect good models to pass the tests and bad models to fail. In the empirical results, the test suite behaved as follows:

\noindent
\textbf{Failed} (as desired) on Problems 20 and 99; 
\textbf{Passed} on Problems 49, 95, 105, 158, and 199 (all correctly modeled, as desired); 
\textbf{Failed} on Problems 9 and 191 despite correct models—these are false positives.

To summarize, our test suite correctly classified 5 models as correct and 2 models as incorrect. No optimization models were falsely classified as correct, while 2 models were falsely classified as incorrect—overall demonstrating good accuracy in this test.

\section{Conclusions and Further Work.}
We introduced an automated, multi-agent framework for validating optimization models that are generated by LLMs from natural-language descriptions. Borrowing techniques from software testing, the framework (i) creates a problem-specific testing API, (ii) automatically composes test cases using this API, and (iii) applies optimization-oriented mutation testing to probe edge cases. Experiments show that the approach achieves strong mutation-coverage scores, demonstrating its effectiveness at uncovering errors and confirming model correctness. As future work, two main directions are of interest: (i) refining the mutation process to achieve higher coverage, and (ii) evaluating the framework on more complex, real-world problems.
%provide intuition why two different modeled provided results that appear in table 1

\bibliography{LLMOptValidation}
\bibliographystyle{unsrt}

\end{document}